\documentclass[3p]{elsarticle}

\usepackage{hyperref}

\journal{Array}

\usepackage{caption} 
\usepackage{array}
\usepackage{graphicx}
\usepackage{amsmath}
\usepackage{amssymb}
\usepackage{multicol}
\usepackage{caption}
\usepackage{comment}
\usepackage{xcolor}

\usepackage{multirow}
\usepackage{tabularx}
\usepackage{adjustbox}
\usepackage{subfig}             
\usepackage{tabularx}

\begin{document}

\begin{frontmatter}

\title{Evaluation of Tree Based Regression over Multiple Linear Regression for Non-normally Distributed Data in Battery Performance \tnoteref{mytitlenote}}








\author[1,4]{Shovan Chowdhury}

\author[2]{Yuxiao Lin}

\author[3]{Boryann Liaw \corref{mycorrespondingauthor1}}
\cortext[mycorrespondingauthor1]{Corresponding author:}
\ead{boryann.liaw@inl.gov}

\author[4]{Leslie Kerby \corref{mycorrespondingauthor1}} 
\ead{kerblesl@isu.edu}

\address[1]{Department of Mechanical Engineering, Idaho State University, Pocatello, Idaho 83209, USA}
\address[2]{Materials Science and Engineering, Energy and Environment Science and Technology,
Idaho National Laboratory, Idaho Falls, Idaho 83415, USA}
\address[3]{Energy Storage and Advanced Transportation,Energy and Environment Science and Technology,
Idaho National Laboratory, Idaho Falls, Idaho 83415, USA}
\address[4]{Computational Engineering and Data Science Laboratory, Department of Computer Science,
Idaho State University, Pocatello, Idaho 83209, USA}










\begin{abstract}
Battery performance datasets are typically non-normal and multicollinear. Extrapolating such datasets for model predictions needs attention to such characteristics. This study explores the impact of data normality in building machine learning models.  In this work, tree-based regression models and multiple linear regressions models are each built from a highly skewed non-normal dataset with multicollinearity and compared. Several techniques are necessary, such as data transformation, to achieve a good multiple linear regression model with this dataset; the most useful techniques are discussed. With these techniques, the best multiple linear regression model achieved an $R^2=81.23\% $ and exhibited no multicollinearity effect for the dataset used in this study. Tree-based models perform better on this dataset, as they are non-parametric, capable of handling complex relationships among variables and not affected by multicollinearity. We show that bagging, in the use of Random Forests, reduces overfitting. Our best tree-based model achieved accuracy of $R^2=97.73\%$. This study explains why tree-based regressions promise as a machine learning model for non-normally distributed, multicollinear data.
\end{abstract}


\end{frontmatter}


\section{Introduction}

In this study, a highly skewed, non-normally distributed, multicollinear dataset that delineates a set of battery cycle life performance under high-rate charging is used. To develop a suitable life prediction model, we studied the feasibility of several regression methods. Through analyses, the suitability of tree-based approach is further explained. We began with two multiple linear regression predictive models: one without data transformation and the other with the Box-Cox transformation. Both utilize stepwise regression to address multicollinearity. The importance of normally distributed variables is examined, and the impact on multicollinearity is explored, from the results of these two models. Two tree-based regression models are then built: a decision tree model and a random forest regression model. For these tree-based models, no data transformation was performed. The suitability of all four models is compared.\\


In the field of data science and machine learning, regression is a process of obtaining correlation between dependent and independent variables. When the response variable is continuous in nature, one can use regression algorithms for developing a predictive model. Among all the regression algorithms, linear regression is the most common and ancient technique. For multiple linear regression (more than one independent variable), some of the assumptions are: i) residuals should be normally distributed, ii) there should be a linear relationship between independent and dependent variables, iii) independent variables are not significantly correlated (i.e., no multicollinearity) and iv) homoscedasticity of errors \cite{Osborne2002}. However, there is significant argument about how important a normal distribution of variables is. Schimdt and Finan \cite{Schmidt2018} concluded in their work that linear regression models are robust to violation of normality assumption of variables when there is a large sample size. Williams, et al. \cite{Williams2013} argues that it is not mandatory to have variables with normal distributions for building regression models. Rather, the authors suggest focusing on other assumptions like normality and equal variance of error and some potential problems such as multicollinearity among the variables, outliers, etc.\\

Multicollinearity occurs when there is a strong correlation between predictors in a dataset \cite{Akhil2013}. The problem of multicollinearity might not affect the accuracy of an MLR model, but multicollinearity makes it difficult to interpret how predictor variables impact the response variable \cite{Ranjit2006}. This problem can be solved by using the variable screening method. Stepwise regression is one of the variable screening methods \cite{Claudio2002} which can reduce negative multicollinearity impacts. Regarding outliers, one can remove the outlier, but this can lead to a biased model. Data transformation is suggested to treat this problem \cite{osborne2002multiple}. When variables are substantially non-normally distributed, data transformation often improves the MLR accuracy \cite{osborne2010improving}. The use of the Box-Cox transformation can reduce both non-normality and outliers in the data \cite{sakia1992box}. \\

In recent years, nonparametric algorithms have been developed to solve regression problems; these algorithms do not have any normality or multicollinearity assumptions and so transformation of the dataset is not required \cite{breiman2001statistical}. One such algorithm, the decision tree, is a tree-like structure consisting of a root node at the top, connecting to layers of intermediate nodes, and ending in a set of terminal nodes (leaves) at the bottom \cite{xu2005decision}. Each node contains a binary decision and layers of nodes continue until some stopping criterion is achieved. This method is usually capable of achieving high accuracy, but it might suffer from overfitting \cite{breiman1984classification}. Random forests are an ensemble learning method consisting of many (often hundreds of) decision trees. It averages the prediction from each decision tree in the ensemble, leading to a reduction in bias and overfitting, and an increase in accuracy \cite{breiman2001random} \cite{cutler2007random}.

\section{Data collection and target analysis}

\begin{figure}[t]
    \centering
    \includegraphics{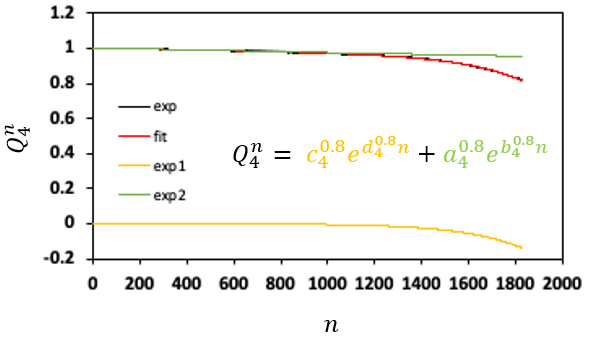}
    \caption{A binomial model describing the fading of the normalized capacity in cycle aging and life.}
    \label{figure1}
\end{figure}

\begin{figure}[t]
  \centering
  \subfloat[$d_i^{0.99}$ vs $\frac{1}{n_i^{0.8}}$]{\includegraphics[width=0.45\textwidth]{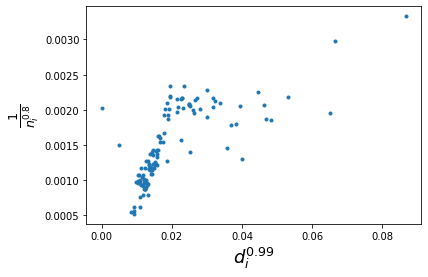}\label{corr1}}
  \hfill
  \subfloat[$d_i^{0.98}$ vs $\frac{1}{n_i^{0.8}}$]{\includegraphics[width=0.45\textwidth]{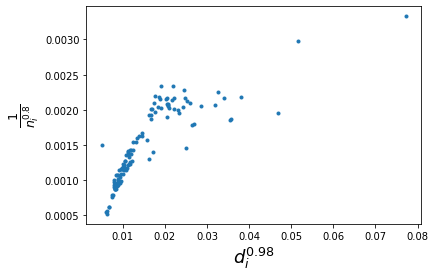}\label{corr2}} \\
  \subfloat[$d_i^{0.97}$ vs $\frac{1}{n_i^{0.8}}$]{\includegraphics[width=0.45\textwidth]{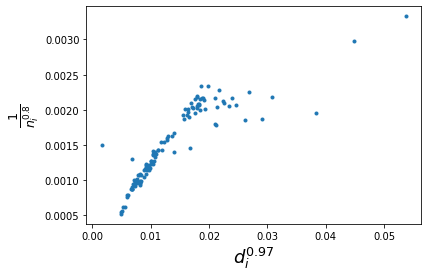}\label{corr3}}
  \hfill
  \subfloat[$d_i^{0.80}$ vs $\frac{1}{n_i^{0.8}}$]{\includegraphics[width=0.45\textwidth]{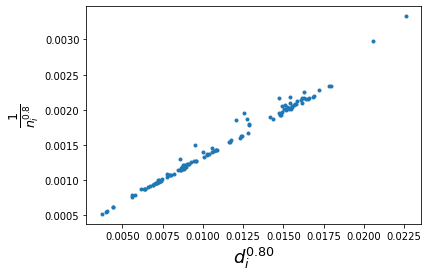}\label{corr4}}
  \caption{Correlation between $\frac{1}{n_i^{0.8}}$ and each variable}
   \label{Corr of each variable}
\end{figure}

The dataset used in this research was the data made available in the publication in \cite{severson2019data}. The dataset consists of 124 commercial lithium iron phosphate batteries (A123, Livonia, Michigan) cycled with a variety of fast charging protocols to reach the end-of-life (EOL) condition defined as $80\%$ of the nominal capacity. Using this dataset, a mechanistic binomial model was developed and used in this work \cite{Lin2021} to provide the mechanistic insight of the dataset regarding the capacity loss attributes and their contributions to the capacity degradation (see Figure-\ref{figure1}) . Table-\ref{parameter definition} outlines the corresponding parameters of this binomial model. From this model, the values of the first exponential term (d) can be obtained for each battery at various points in its cycle life: when discharge capacity dropped to $99\%$, $98\%$, $97\%$ and $80\%$ of its initial nominal value. A small portion of the data is shown in Table-\ref{sample dataset}  in descending order from a total of 124 observations. Here, the exponential term $d$ represents the extent of loss of active material (LAM) in the binomial model at different points in its cycle life. Four different $d$ values are presented when the capacity retention $(Q)$ reached $99\%$, $98\%$, $97\%$ and $80\%$ of its initial nominal value, respectively. Correlation between the value of 1/(cycle number at EOL or $n_i^Q$) and d value at different stages of cycle life is given in Figure-\ref{Corr of each variable}. There is a strong correlation between $d_i^{0.8}$ and $\frac{1}{n_i^{0.8}}$. Our goal is to predict the cycle life from the early cycle $d$ value, i.e., $d_i^{0.99}$, $d_i^{0.98}$ $\&$ $d_i^{0.97}$. As there is a good linear relationship between $d_i^{0.8}$ and $n_i^{0.8}$, predicting the cycle life from the early cycle $d$ values shall depend on how well the conformity of these early cycle $d$ values to such a linear relationship is, so the accuracy of the life prediction can be further assessed. In this work, multiple linear regression and tree-based regression algorithms were utilized for this assessment. Comparison of the results from different regression methods shall help us understand any limitation of linear regression over random forest regression and importance of data transformation while data is not normally distributed. For the highly skewed non-normal dataset used in this study, tree-based models outperform linear regression models in terms of accuracy and robustness. Normal distribution of variables is turned out to be an important factor for getting a good multiple linear regression model, whereas it is not necessary for the tree-based models. Using Box-Cox transformation for the dataset increases the prediction accuracy.

\begin{table}[t]
 \caption{Parameter Definition of the Binomial Model}
 \vspace{2pt}
    \centering
    \begin{tabular}{ccm{11cm}}
    \hline
        Parameter & Value & Physical Meaning \\ \hline
        $n$ & & Cycle number\\
        $i$ & & Cell number\\
        $Q_i^n$ & & Capacity of cell $i$ at cycle $n$, normalized by the initial capacity\\
        $n_i^Q$ & & Cycle number when the normalized capacity of cell $i$ drops below $Q$\\
        $a_i^Q$ & $1.003$ & Initial $Q_i^n$, close to 1 at the beginning of life\\
        $b_i^Q$ & $-2.67 \times 10^{-5}$ & Parameter related to the capacity loss due to LLI of cell i, obtained from the initial cycle to cycle $n_i^Q$ \\
        $c_i^Q$ & $-8.92 \times 10^{-5}$ & Initial loss due to LAM, close to 0 at the beginning of life \\
        $d_i^Q$ & $4.02 \times 10^{-3}$ & Parameter related to the capacity loss due to LAM of cell i, obtained from the initial cycle to cycle $n_i^Q$\\
        $R^2$ & $9.99 \times 10^{-1}$ & Correlation Coefficient\\ \hline
    \end{tabular}
    \label{parameter definition}
\end{table}

\begin{table}[!tbp]
 \caption{Sample Dataset}
 \vspace{2pt}
    \centering
    \begin{tabular}{cccccc}
    \hline
        $n_i^{0.8}$ & $\frac{1}{n_i^{0.8}}$ & $d_i^{0.99}$ & $d_i^{0.98}$ & $d_i^{0.97}$ & $d_i^{0.80}$ \\ \hline
        1935 & 0.0005168 & 0.009260363 & 0.006162207 & 0.004907865 & 0.003719588 \\
        1836 & 0.00054466 &	0.008406241 & 0.005927148 & 0.004872629 & 0.003975488\\
        1801 & 0.00055525 & 0.009136795 & 0.006179498 & 0.005038435 & 0.004043477 \\
        1642 & 0.00060901 & 0.009246823 & 0.006631469 & 0.005672362 & 0.004429832 \\ \hline
    \end{tabular}
    \label{sample dataset}
\end{table}

\section{Machine Learning Algorithm}

Three machine learning techniques are used in this work: (1) a traditional multiple linear regression with stepwise technique, (2) tree-based regressions like decision tree, and (3) ensemble learning method such as random forest regression is used.

\subsection{Multiple Linear Regression}

Multiple Linear Regression (MLR) is a popular statistical technique to find out the relationship between predictor variables and response variables. In this method, a linear relationship is modeled between independent variable (predictors) and dependent variable (response). This is like the ordinary least square (OLS) method, but multiple variables can be used in place of a single variable. The formula for MLR is,

\begin{equation}\label{MLR}
    y_i = \beta_0 + \beta_1 x_1 + \hdots + \beta_p x_{ip} + \epsilon_i,
    \end{equation}

Where, $\hat{y}_i$ is the response variable, $x_i$ are predictors, $\beta_0$ is constant coefficient and $\beta_p$ are independent variables coefficient. There is an error term $\epsilon_i$ which add noise to the model. For minimizing this random error, sum of square error (SSE) function is used,

\begin{equation}
    \textrm{SSE} = \sum\limits_{i=1}^n (y_i - \hat{y}_i)^2
\end{equation}

where $y_i$ is the observed value and $\hat{y}_i$ the predicted value from the model. The best fitted model will give the lowest SSE value. Among independent variables, there should not be high correlation as in that case the same information will be presented by highly correlated variables to the model. For this reason, one can use stepwise regression for mitigating multicollinearity problems among variables and making the model less complex. Another assumption of MLR is that residuals should be independent and normally distributed with mean of 0 and variance of $\sigma$. One can use Durbin statistics for determining the correlation between residuals \cite{hill1987use}. However, the size of the error in the prediction also shouldn’t change significantly across the values of the independent variable. 


\begin{figure}[t]
    \centering
    \includegraphics[width= 13cm]{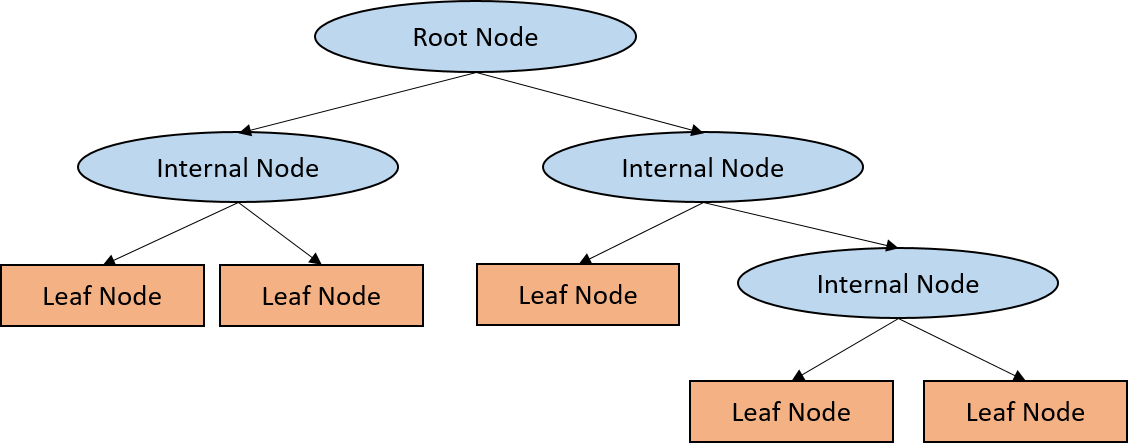}
    \caption{Decision Tree structure}
    \label{fig:dt}
\end{figure}

\begin{figure}[!tbp]
    \centering
    \includegraphics[width= 11cm]{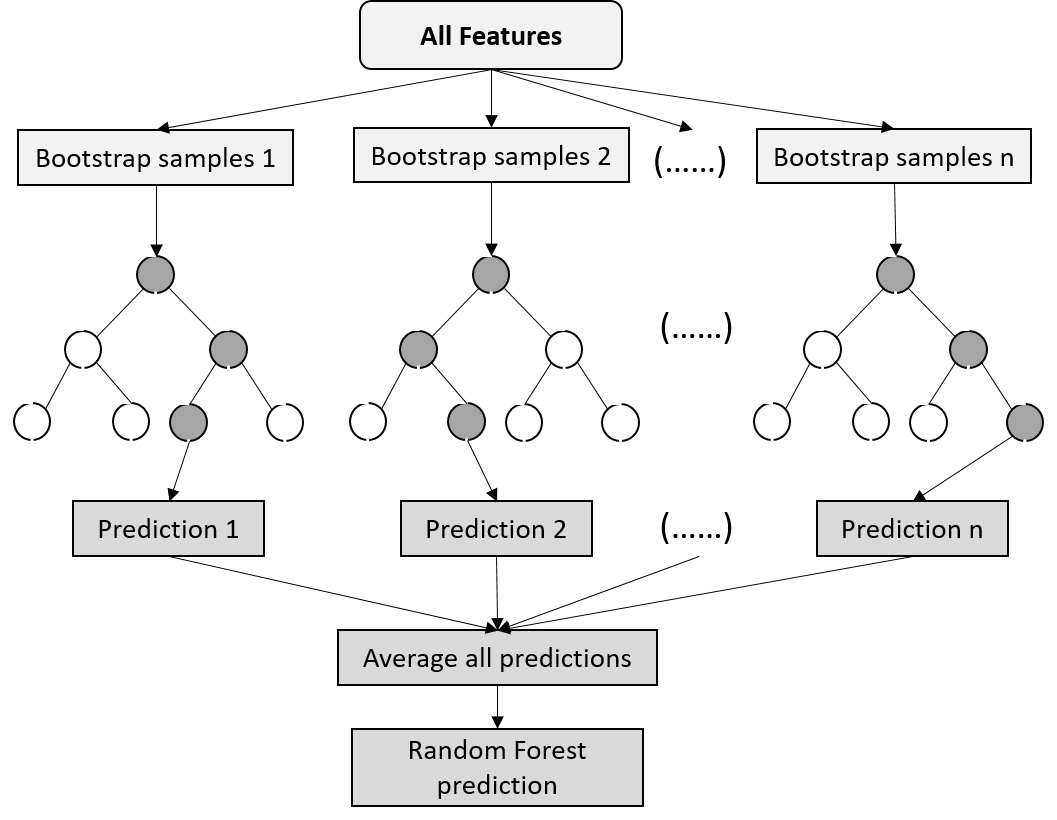}
    \caption{Random Forest Regression Algorithm}
    \label{fig:rfr}
\end{figure}
\subsection{Decision Tree}

The Decision tree algorithm is a non-parametric supervised machine learning method that is used for both classification and regression. It is a tree like structure consists of several branch (see Figure-\ref{fig:dt}), where each internal node denotes a test on an attribute, each branch represents an outcome of the test, and each leaf node (terminal node) holds a predicted value \cite{chowdhury2020research}. Among four types of decision tree algorithm: Iterative Dichotomiser (ID3), Classification and Regression Trees (CART), Chi-Square and Reduction in Variance, the CART algorithm is being widely used for regression problems. In CART regression, data is being split into two groups by finding the threshold that gives the smallest sum of square residual. This procedure is repeated for further splitting until some stopping criterion reached. If any stopping criterion doesn’t set up, then the model will fit the training data perfectly, it probably means it is overfit and will not perform well with new data. In that case, the model will have low bias, but potentially high variance. For preventing the model from overfitting, there are some mitigation techniques. One of the techniques is to split observation when there is some minimum number of samples remaining. Another way, one can set the maximum depth of the tree.

\subsection{Random Forest Regression (RFR)}

As discussed earlier, decision trees might suffer from high variance. To reduce this weakness, random forest regression (RFR) is introduced which constructs many decision trees in one model. This combining technique is called bootstrap aggregation or bagging, as shown in Figure-\ref{fig:rfr}, to reduce the variance of the model. In this method, the same dataset is not used for all decision trees. A separate bootstrap sample (with replacement) from the original dataset is used for building each tree and then average their result to find out the prediction. To minimize the effect of high collinearity among the trees in a forest, RFR uses a subset of features in each decision tree. For selecting m number of subset features from n total features, one rule is $m = \sqrt{n}$. Due to the splitting of predictors, strong predictors might not be able to dominate all the time which reduces overfitting. RFR model don’t need separate cross validation procedures for determining the model performance because it uses out-of-bag (OOB) samples to validate the model built with training samples. \cite{ouedraogo2019application} \cite{vceh2018estimating}

\section{Methodology}

\begin{figure}[!tbp]
    \centering
    \includegraphics[width= 13cm]{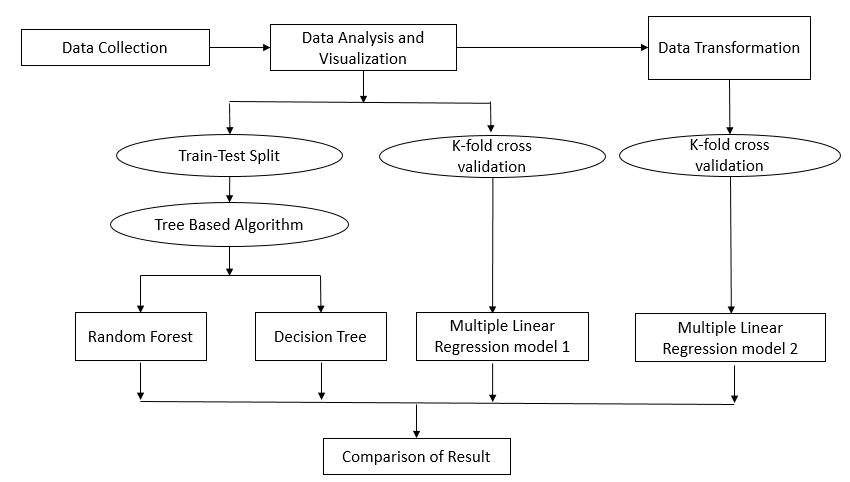}
    \caption{Flowchart of the whole method}
    \label{fig:flowchart}
\end{figure}

The scope of the present study is to compare and understand the performance of MLR-based and tree-based models on this non-normal dataset. The entire process of work is shown in Figure-\ref{fig:flowchart}. First, data visualization is done to find whether our features are normally distributed or not. Then for one of the MLR models, data is transformed using Box-Cox transformation. For MLR models, k-fold cross validation is used to determine the model validation. One MLR model is developed using transformed data and another with the original data. Effect of non-normality in the dataset is scrutinized by observing the performance of these two models. In the RFR and decision tree model, 80 percent of the data is used for training and the rest of the data is used for testing. Finally, the accuracy of the different algorithms is compared using some resulting parameter.

\subsection{Data Visualization}

\begin{figure}[!tbp]
  \centering
  \subfloat[Statistics for $d_i^{0.99}$]{\includegraphics[width=0.45\textwidth]{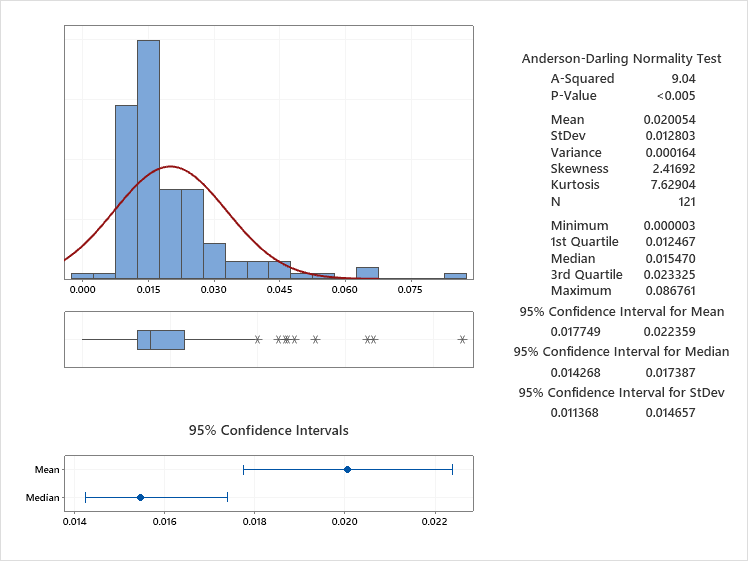}\label{fig:f1}}
  \hfill
  \subfloat[Statistics for $d_i^{0.98}$]{\includegraphics[width=0.45\textwidth]{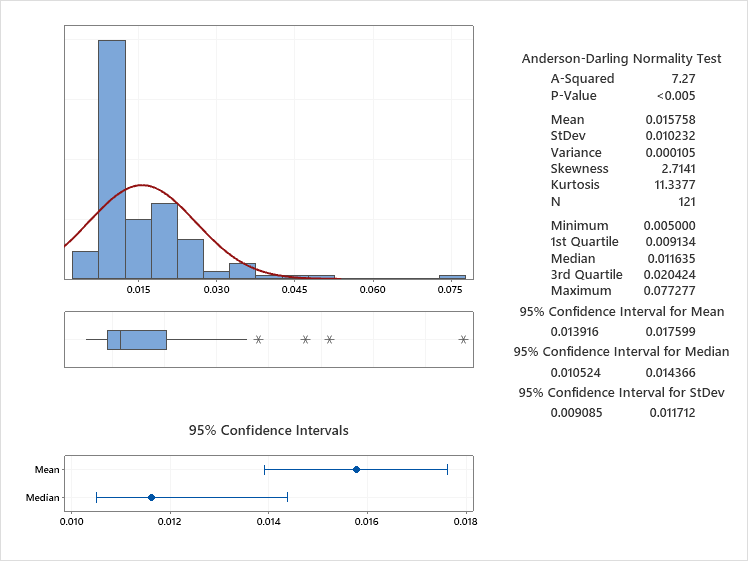}\label{fig:f2}} \\
  \subfloat[Statistics for $d_i^{0.97}$]{\includegraphics[width=0.45\textwidth]{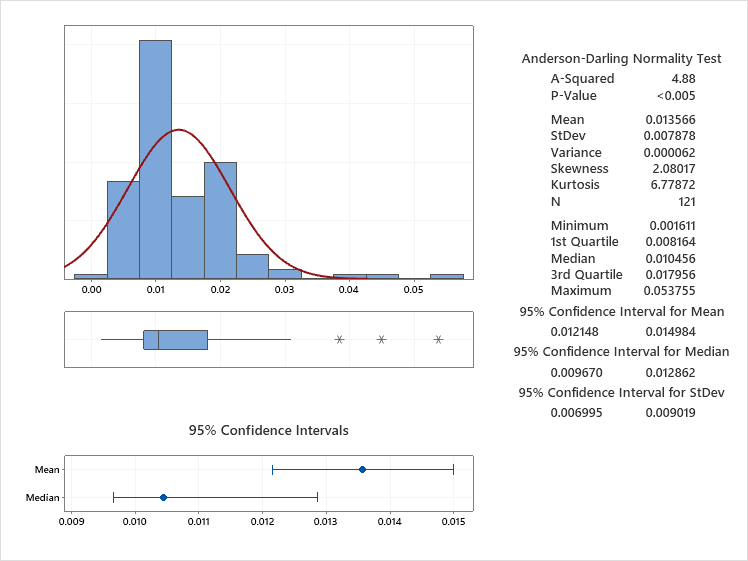}\label{fig:f3}}
  \hfill
  \subfloat[Statistics for $d_i^{0.80}$]{\includegraphics[width=0.45\textwidth]{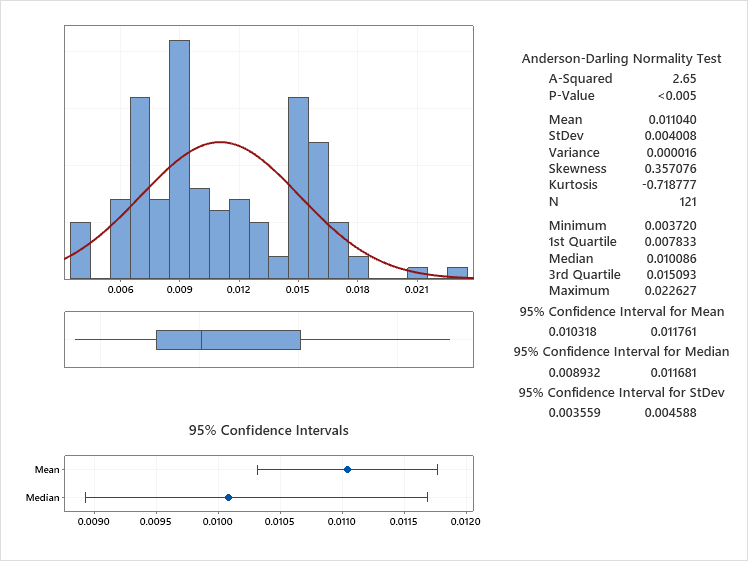}\label{fig:f4}}
  \caption{Histogram and descriptive statistics of features}
\label{fig:summary report before}
\end{figure}

When skewness of a variable falls between -0.80 to 0.80, it may be considered as normally distributed. Histograms of the independent and dependent features are presented in Figure-\ref{fig:summary report before}. It is clearly observed from the histogram that none of the features is normally distributed. All the independent features are right skewed. There are some outliers observed in the data histogram. Correlation between each independent feature is presented in TABLE III. High correlation (0.98) between $d_i^{0.98}$ and $d_i^{0.97}$ is detected, meaning the data suffers from multicollinearity. 

\subsection{Data Transformation}

\begin{table}[!tbp]
 \caption{CORRELATION MATRIX OF INDEPENDENT FEATURES}
  \vspace{2pt}
    \centering
    \begin{tabular}{c|ccc}
    
         & $d_i^{0.99}$ & $d_i^{0.98}$ & $d_i^{0.97}$ \\ \hline
        $d_i^{0.99}$ & 1 & 0.98 & 0.90 \\
        $d_i^{0.98}$ & 0.93 & 1 & 0.98 \\
        $d_i^{0.97}$ & 0.90 & 0.98 & 1 \\ \hline
    \end{tabular}
    \label{table:PERFORMANCE OF RFR MODEL}
\end{table}

\begin{figure}
    \centering
    \includegraphics[width= 10cm]{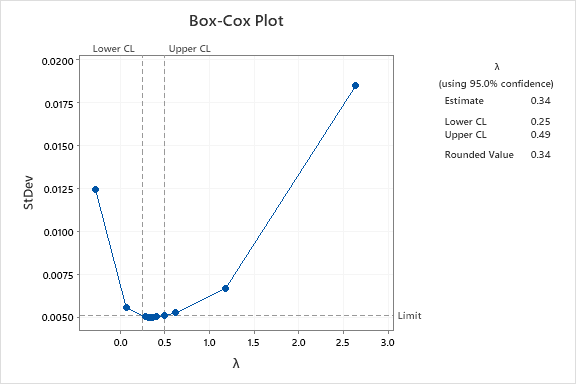}
    \caption{Box-cox transformation of $d_i^{0.99}$}
    \label{fig:boxcox}
\end{figure}

\begin{figure}[t]
  \centering
  \subfloat[Statistics for $d_i^{0.99}$ after transformation]{\includegraphics[width=0.45\textwidth]{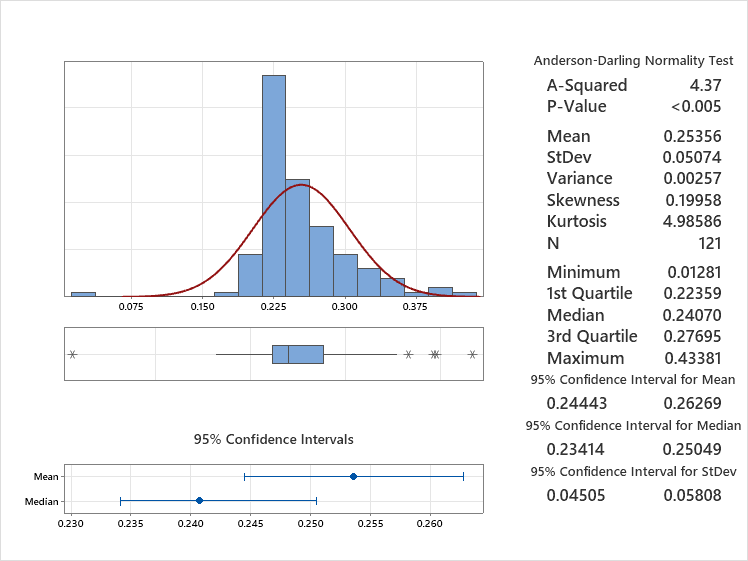}\label{fig:f1t}}
  \hfill
  \subfloat[Statistics for $d_i^{0.98}$ after transformation]{\includegraphics[width=0.45\textwidth]{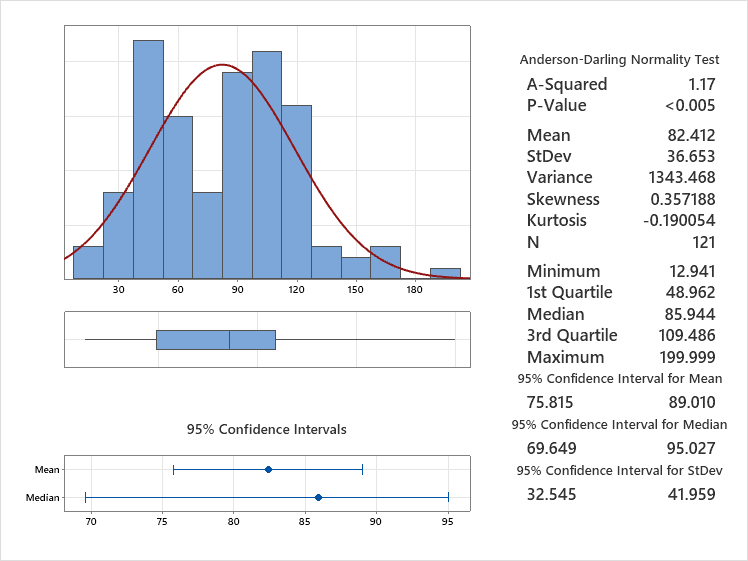}\label{fig:f2t}} \\
  \subfloat[Statistics for $d_i^{0.97}$ after transformation]{\includegraphics[width=0.45\textwidth]{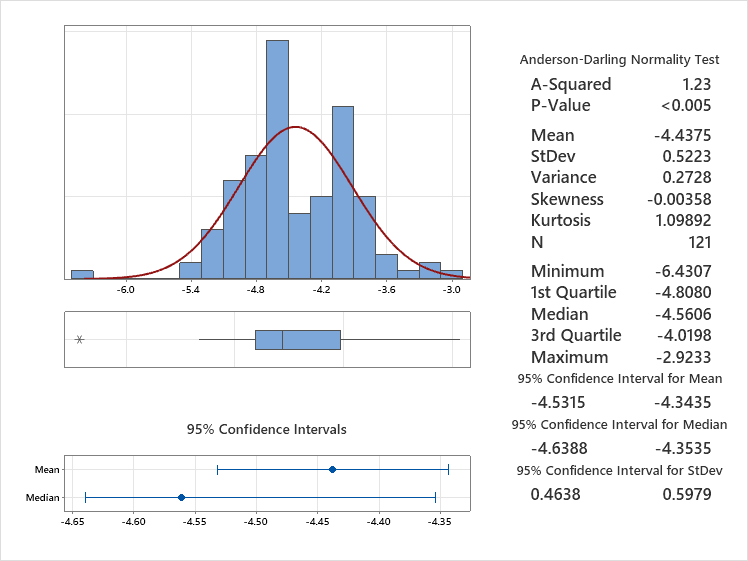}\label{fig:f3t}}
  \hfill
  \subfloat[Statistics for $d_i^{0.80}$ after transformation]{\includegraphics[width=0.45\textwidth]{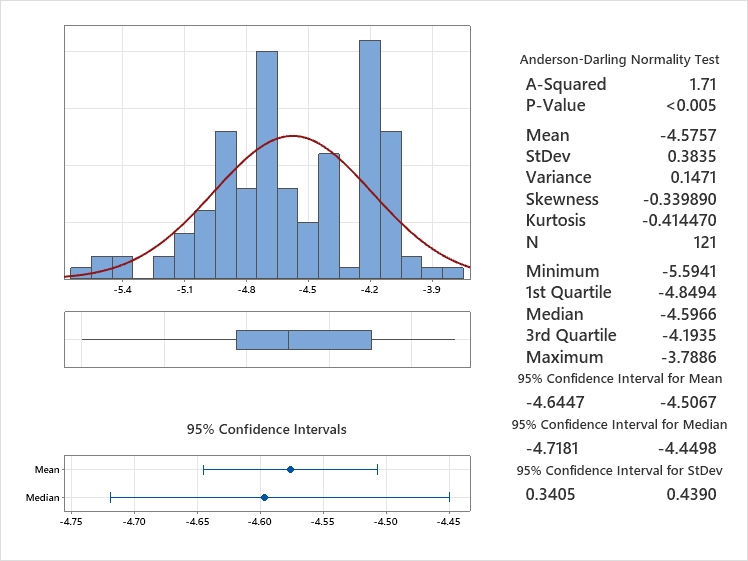})\label{fig:f4t}}
  \caption{Histogram and descriptive statistics of features after transformation}
  \label{fig:summary report after}
\end{figure}

\begin{figure}[t]
  \centering
  \subfloat[Correlation increased for feature $d_i^{0.98}$ after transformation]{\includegraphics[width=0.45\textwidth]{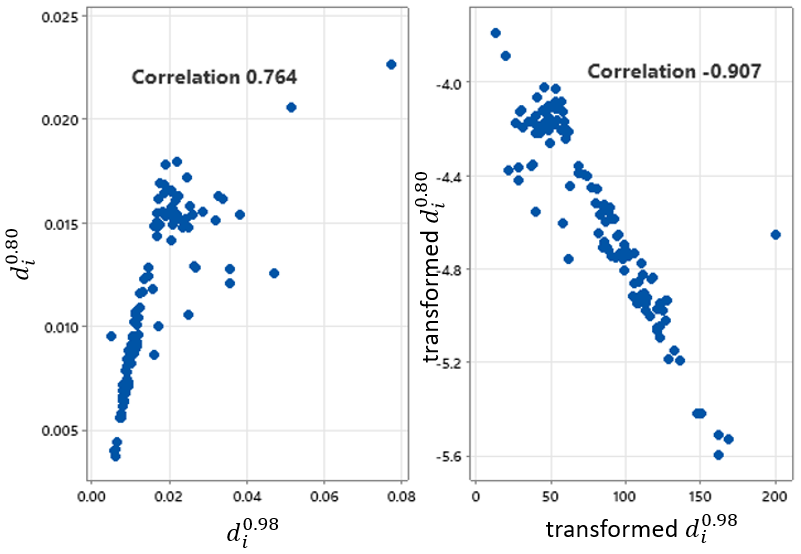}\label{fig:cor1}}
  \hfill
  \subfloat[Correlation increased for feature $d_i^{0.97}$ after transformation]{\includegraphics[width=0.45\textwidth]{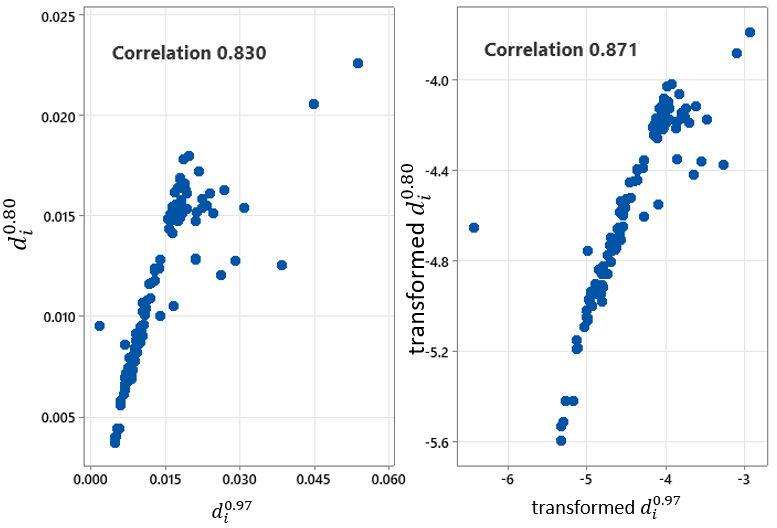}\label{fig:cor2}} \\
  \subfloat[Correlation decreased for feature $d_i^{0.99}$ after transformation]{\includegraphics[width=0.45\textwidth]{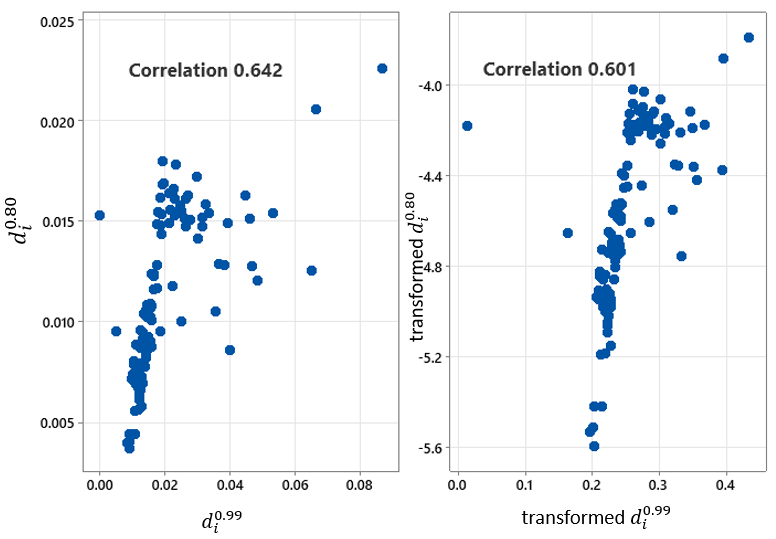}\label{fig:cor3}}
  \caption{Correlation between independent and dependent features before and after transformation}
  \label{fig:correlation change}
\end{figure}

When data is not normally distributed, data transformation can be used to reduce skewness and make it more normal. There are various types of transformation techniques: log transformation, inverse transformation, Box-Cox transformation, etc. In this work, Box-Cox transformation is utilized for transforming the features. The mathematical formula for Box-Cox transformation is given below:
\begin{equation}
  y(\lambda)=\begin{cases}
    \frac{y^{\lambda-1}}{\lambda}, & \text{if $\lambda \not= 0$}.\\
    \log{y}, & \text{if $\lambda = 0$}.
  \end{cases}
\end{equation}

The lambda values usually vary from -5 to 5 and the optimal lambda value is one which represents the best skewness of the distribution. When the value of lambda is zero, then log transformation will be used. Figure-\ref{fig:boxcox} shows how optimum lambda value is calculated for $d_i^{0.99}$. Figure-\ref{fig:summary report after} represents the histogram and descriptive statistics of transformed features which show that the value of skewness is reduced and the data is more normal. Ideally, skewness should be closer to 0. The skewness of independent variable $d_i^{0.99}$ is reduced from 2.41 to 0.19, $d_i^{0.98}$ is reduced from 2.71 to 0.35, $d_i^{0.97}$ is reduced from 2.08 to -0.003 and dependent feature $d_i^{0.80}$ is reduced from 0.35 to -0.33. All the transformed features have skewness between -0.80 to 0.80 which confirms that the transformed features are almost normal. The linear relationship between the predictor and response variable is very important for linear regression. The higher the correlation between predictor and response variable, the higher the model accuracy. Here, correlation with the response variable is increased for the $d_i^{0.97}$ and $d_i^{0.98}$ due to the transformation of the data represented in Figure-\ref{fig:cor1} \& \ref{fig:cor2}. But for the $d_i^{0.99}$, correlation decreased slightly (see Figure-\ref{fig:cor3}). The most significant change is observed for transformed feature $d_i^{0.98}$: it shows an almost perfect linear relationship with transformed $d_i^{0.80}$, with the exception of an outlier. 

\subsection{Model Building}

Minitab is utilized for building the MLR models. Two MLR models are developed for observing the effect of non-normality in the model: one model uses an original dataset with non-normal variables, the second model uses a transformed dataset with more normal variables. Stepwise regression is utilized for each of the MLR models, where the choice of predictive variables is carried out by an automatic procedure which objectively determines which independent variables are the most important predictors for the model. This method selects variables with the help of t-test value. This procedure is continued until no further independent variables can be found that yield significant t-values (at the specified $\alpha$ level) in the presence of the variables already in the model. This is also called variable screening procedure. In our models, we set the value of $\alpha$ at 0.15. K-fold cross validation is utilized with 10 folds. Lastly, residual analysis is done to observe whether the residuals are normally distributed and have equal variance or not.\\

The sklearn python environment is used for developing the tree-based models. The decision tree and RFR models are non-parametric, meaning they are capable of handling non-linear effects. For that reason, the original dataset (without transformation) is used for the tree-based models. As we discussed before, cross validation is not required for the RFR model. This RFR model is built with 80\% of the data being used in training and the remaining 20\% set aside for testing. The RFR is built with 100 decision trees where each tree uses a bootstrap sample from the original dataset. Then, feature importance is obtained to illuminate which feature(s) play a significant role in prediction.   

\section{Results}

Results are evaluated by the coefficient of determination, root mean squared error, mean absolute percentage error and the correlation coefficient.

\begin{itemize}
  \item Coefficient of Determination: The coefficient of determination ($R^2$) is the proportion of the variance in response variable that is explained by the model. It is said to be an accuracy of the regression model. The formula:
  \begin{equation}
    R^2 = 1-\frac{SS_{res}}{SS_{tot}}
  \end{equation}
  where, $SS_{tot}=\sum\limits_{i=1}^n (y_i - \bar{y})^2$  is the total sum of squares and  $SS_{res}=\sum\limits_{i=1}^n (y_i - p_i)^2$ the sum of squares of residuals. Here, $p_i$ is the predicted value from the model.
  \item RMSE: The root-mean-square error (RMSE) is a measure of the differences between values predicted by a model and the observed values.
  \begin{equation}
   \textrm{RMSE} = \sqrt{\frac{\sum\limits_{i=1}^n (p_i-x_i)^2}{N}}
  \end{equation}
  where $p_i$ is the predicted value and $x_i$ is the original value for N number of observations.
  \item MAPE: The mean absolute percentage error (MAPE) is a measure of prediction accuracy of a regression method in statistics. It expresses the accuracy by the following formula:
  \begin{equation}
   \textrm{MAPE} = \frac{1}{n}\sum\limits_{i=1}^n|\frac{A_i-P_i}{A_i}|
  \end{equation}
  where $A_i$ is the actual value, $P_i$ the predicted value and n the number of observations. 
  \item Correlation coefficient: A correlation coefficient is a numerical measurement of the strength of linear relationship between two variables.
\end{itemize}

\subsection{Result of MLR model 1 without transformation}


\begin{table}[t]
     \caption{Regression Summary for MLR 1}
\begin{minipage}{\dimexpr.50\linewidth-2em}
\scriptsize

    \begin{tabular}{{m{0.08\linewidth}m{0.1\linewidth}m{0.05\linewidth} m{0.11\linewidth}m{0.05\linewidth}m{0.11\linewidth}m{0.05\linewidth}}}
    \multicolumn{7}{l}{\footnotesize \textcolor{cyan}{\textbf{Stepwise Selection of Terms}}}\\
         \multicolumn{7}{l}{}\\
     \multicolumn{7}{l}{\footnotesize \textbf{Candidate terms:} $d_i^{0.99}$, $d_i^{0.98}$, $d_i^{0.97}$ }\\
          \multicolumn{7}{l}{}\\
 & \multicolumn{2}{c}{\textbf{---Step 1---}} &\multicolumn{2}{c}{\textbf{---Step 2---}} & \multicolumn{2}{c}{\textbf{---Step 3---}} \\ 
 & \textbf{Coef} & \textbf{P} & \textbf{Coef} & \textbf{P} & \textbf{Coef} &  \textbf{P} \\ \hline
   {\tiny Constant} & 0.0053 &  & 0.0044 &  & 0.0047 &  \\
    $d_i^{0.97}$ & 0.4222 & 0.00 & 1.125 & 0.00 & 1.067 &  0.00 \\
    $d_i^{0.98}$ &  &  & -0.551 & 0.00 & -0.367 &  0.001 \\
    $d_i^{0.99}$ &  &  &  &  & -0.120 &  0.002 \\
    \multicolumn{7}{l}{}\\
 S & \multicolumn{2}{c}{0.00224} &\multicolumn{2}{c}{0.00197} & \multicolumn{2}{c}{0.00190} \\
 R-sq & \multicolumn{2}{c}{68.87\%} &\multicolumn{2}{c}{76.06\%} & \multicolumn{2}{c}{77.87\%} \\
 R-sq(adj) & \multicolumn{2}{c}{68.61\%} &\multicolumn{2}{c}{75.65\%} & \multicolumn{2}{c}{77.30\%} \\
 Mallows $C_p$ & \multicolumn{2}{c}{47.60} &\multicolumn{2}{c}{11.60} & \multicolumn{2}{c}{4.00} \\
 AICc & \multicolumn{2}{c}{-1128.31} &\multicolumn{2}{c}{-1157.93} & \multicolumn{2}{c}{-1165.30} \\
 BIC & \multicolumn{2}{c}{-1120.13} &\multicolumn{2}{c}{-1147.09} & \multicolumn{2}{c}{-1151.84} \\
     \multicolumn{7}{l}{}\\
\multicolumn{7}{l}{  $\alpha$ to enter = 0.15, $\alpha$ to remove = 0.15}\\
    \end{tabular}
    \label{table4}
    \end{minipage}
    \begin{minipage}{\dimexpr.48\linewidth-2em}
\scriptsize
    \centering
    \begin{tabular}{m{0.08\linewidth}m{0.09\linewidth}m{0.09\linewidth} m{0.13\linewidth}m{0.12\linewidth}m{0.14\linewidth}m{0.14\linewidth}}
    \multicolumn{7}{l}{\footnotesize\textcolor{cyan}{ \textbf{Model Summary}}}\\
         \multicolumn{7}{l}{}\\
        \textbf{S} & \textbf{R-sq} & \textbf{R-sq (adj)} & \textbf{PRESS} & \textbf{R-sq (pred)} & \textbf{10-fold S} &  \textbf{10-fold R-sq} \\ \hline
        0.0019 & 77.87\% & 77.30\% & 0.0005386 & 72.06\% & 0.0021673 &  70.52\%
    \end{tabular}
    \label{table4b}
    \vspace{10pt}
    \begin{tabular}{{m{0.1\linewidth}m{0.125\linewidth}m{0.12\linewidth} m{0.21\linewidth}m{0.085\linewidth}m{0.075\linewidth}m{0.075\linewidth}}}
        \multicolumn{7}{l}{\footnotesize\textcolor{cyan}{ \textbf{Coefficients}}}\\
             \multicolumn{7}{l}{}\\
\textbf{Term} & \textbf{Coef} & \textbf{SE Coef} & \textbf{95\% CI} & \textbf{T-Value} & \textbf{P-Value} &  \textbf{VIF}\\ \hline
{ Constant} & 0.004756 & 0.000385 & (0.00399, 0.00551) & 12.37 & 0.000 &  \\
$d_i^{0.99}$ & -0.1202 & 0.0388 & (-0.1970, -0.0434) & -3.10 & 0.002 & 8.11 \\
$d_i^{0.98}$ & -0.367 & 0.107 & (-0.579, -0.154) & -3.41 & 0.001 & 39.73 \\
$d_i^{0.97}$ & 1.067 & 0.118 & (0.834, 1.300) & 9.07 & 0.000 & 28.24 \\
    \end{tabular}
    \label{table4c}
        \vspace{13pt}
    \end{minipage}
    \label{table:MLR 1}
\end{table}

Stepwise regression with a forward selection method is adopted to build this model. Table-\ref{table:MLR 1} shows how an independent variable is selected in each step. At the end of three steps, adjusted $R^2$ value becomes 77.30\%. However, from the model summary in Table-\ref{table:MLR 1}, 10-fold cross validation $R^2$ score is observed as 70.52\% which concludes that the model is overfit. Multicollinearity is determined by variance inflation factor (VIF) and values greater than 10 suffer from severe multicollinearity. From the coefficients in Table-\ref{table:MLR 1}, we see that two of the variables have VIF greater than 10 and hence there is a serious multicollinearity problem. Although p-values are all less than 0.05, and model accuracy is decent, this model suffers from overfitting and multicollinearity.

\subsection{Result of MLR model 2 with transformation}

\begin{table}[t]
      \caption{Regression Summary for MLR 2}
\begin{minipage}{\dimexpr.45\linewidth-2em}
\scriptsize
    \begin{tabular}{m{0.15\linewidth}m{0.17\linewidth}m{0.07\linewidth} m{0.14\linewidth}m{0.07\linewidth}}
    \multicolumn{5}{l}{\footnotesize \textcolor{cyan}{\textbf{Stepwise Selection of Terms}}}\\
         \multicolumn{5}{l}{}\\
     \multicolumn{5}{l}{\footnotesize \textbf{Candidate terms:} $d_i^{0.99}$, transformed $d_i^{0.98}$, }\\  
      \multicolumn{5}{l}{\footnotesize transformed $d_i^{0.97}$ }\\
               \multicolumn{5}{l}{}\\
 & \multicolumn{2}{c}{\textbf{---Step 1---}} &\multicolumn{2}{c}{\textbf{---Step 2---}} \\ 
 & \textbf{Coef} & \textbf{P} & \textbf{Coef} & \textbf{P} \\ \hline
    Constant & -3.7935 &  & -3.5999 &  \\
    transformed $d_i^{0.98}$  & -0.00949 & 0.000 & -0.0107 &  0.000 \\
    $d_i^{0.99}$ & & & -4.68 &  0.006 \\
    \multicolumn{5}{l}{}\\
 S & \multicolumn{2}{c}{0.162076} &\multicolumn{2}{c}{0.157588}\\
 R-sq & \multicolumn{2}{c}{82.29\%} &\multicolumn{2}{c}{83.39\%}\\
 R-sq(adj) & \multicolumn{2}{c}{82.14\%} &\multicolumn{2}{c}{83.11\%} \\
 Mallows $C_p$ & \multicolumn{2}{c}{8.30} &\multicolumn{2}{c}{2.46} \\
 AICc & \multicolumn{2}{c}{-92.79} &\multicolumn{2}{c}{-98.47} \\
 BIC & \multicolumn{2}{c}{-84.61} &\multicolumn{2}{c}{-87.63} \\
     \multicolumn{5}{l}{}\\
\multicolumn{5}{l}{  $\alpha$ to enter = 0.15, $\alpha$ to remove = 0.15}\\
    \end{tabular}
    \label{table4}
    \vspace{35pt}
    \end{minipage}
\begin{minipage}{\dimexpr.50\linewidth-2em}
    \scriptsize
    \centering
       \begin{tabular}{m{0.1\linewidth}m{0.1\linewidth}m{0.12\linewidth} m{0.1\linewidth}m{0.14\linewidth}m{0.145\linewidth}m{0.14\linewidth}}
    \multicolumn{7}{l}{\footnotesize \textcolor{cyan}{\textbf{Model Summary}}}\\
         \multicolumn{7}{l}{}\\
        \textbf{S} & \textbf{R-sq} & \textbf{R-sq (adj)} & \textbf{PRESS} & \textbf{R-sq (pred)} & \textbf{10-fold S} &  \textbf{10-fold R-sq} \\ \hline
        0.15788 & 83.39\% & 83.11\% & 3.54007 & 79.94\% & 0.165785 &  81.15\%
    \end{tabular}
    \label{table4b}
       \vspace{10pt}
    \begin{tabular}{{m{0.15\linewidth}m{0.1\linewidth}m{0.12\linewidth} m{0.25\linewidth}m{0.085\linewidth}m{0.075\linewidth}m{0.075\linewidth}}}
        \multicolumn{7}{l}{\footnotesize \textcolor{cyan}{\textbf{Coefficients}}}\\
             \multicolumn{7}{l}{}\\
\textbf{Term} & \textbf{Coef} & \textbf{SE Coef} & \textbf{95\% CI} & \textbf{T-Value} & \textbf{P-Value} &  \textbf{VIF}\\ \hline
Constant & -3.59 & 0.0776 & (-3.75,-3.44) & -46.4 & 0.000 &  \\
$d_i^{0.99}$ & -4.68 & 1.67 & (-7.99,-1.38) & -2.8 & 0.006 & 2.21 \\
 transformed $d_i^{0.98}$ & -0.01 & 0.00058 & (-0.011,-0.009) & -18.3 & 0.000 & 2021 \\
    \end{tabular}
    \label{table4c}
        \vspace{10pt}
    \begin{tabular}{{m{0.16\linewidth}m{0.04\linewidth}m{0.075\linewidth} m{0.20\linewidth}m{0.08\linewidth}m{0.08\linewidth}m{0.08\linewidth}m{0.075\linewidth}}}
        \multicolumn{8}{l}{\footnotesize \textcolor{cyan}{\textbf{Analysis of Variance}}}\\
             \multicolumn{8}{l}{}\\
\textbf{Source} & \textbf{DF} & \textbf{Seq SS} & \textbf{Contribution} & \textbf{Adj SS} & \textbf{Adj MS} &  \textbf{F-Value} & \textbf{P-Value}\\ \hline
Regression & 2 & 14.717 & 83.39\% & 14.7167 & 7.35837 & 296.30 & 0.000 \\
$d_i^{0.99}$ & 1 & 6.354 & 36.01\% & 0.1955 & 0.19552 & 7.87 & 0.006 \\
 transformed $d_i^{0.98}$ & 1 & 8.363 & 47.39\% & 8.3627 & 8.36266 & 336.74 & 0.000 \\
 Error & 118 & 2.930 & 16.61\% & 2.9304 & 0.02843 & & \\
 Total & 120 & 17.647 & 100.00\% & & & &  \\
    \end{tabular}
    \label{table4c}
            \vspace{1pt}
    \end{minipage}
        \label{table:MLR 2}
\end{table}

\begin{figure}[!tbp]
    \centering
    \includegraphics[width= 10cm]{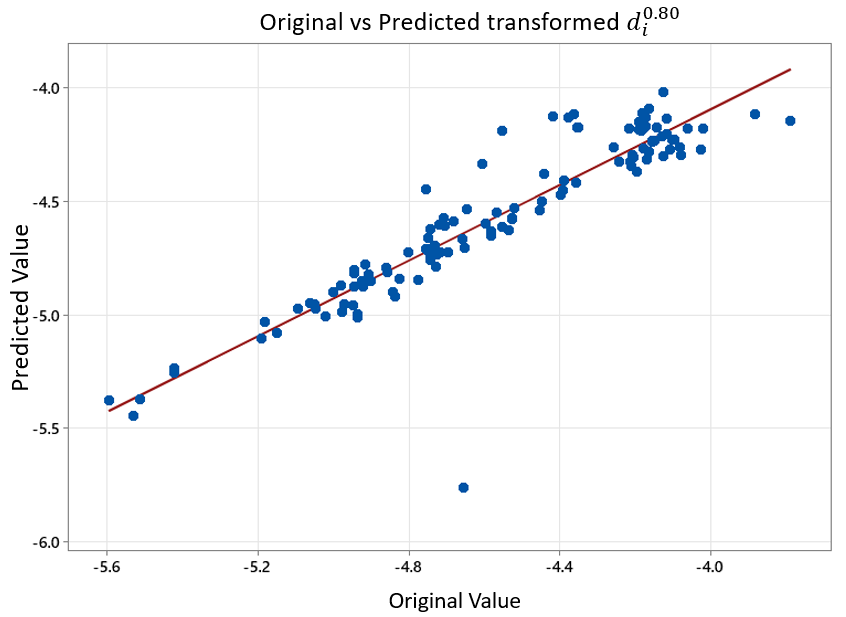}
    \caption{Original vs Predicted Observation in MLR model 2}
    \label{fig:Original vs Predicted Observation in MLR model 2}
\end{figure}

After trying numerous models with different features, the best model from MLR is found by using transformed features (except for $d_i^{0.99}$) and stepwise regression. The results presented in Table-\ref{table:MLR 2} show how stepwise regression found that the transformed $d_i^{0.98}$ is the most important independent variable able to explain 82.29\% variation of the model by itself, followed by $d_i^{0.99}$ which when added increases the $R^2$ score to 83.39\%. Stepwise regression eliminates one feature $d_i^{0.97}$ and makes the model less complicated. It reaches Mallow’s $C_p$ of 2.46 in step 2 which is almost equal to the number of predictors and good enough to stop the regression in that step. The regression equation for this model is,
\begin{equation}
   \texttt{transformed}\_d_i^{0.80} = -3.5999-4.68[d_i^{0.99}]- 0.010701[ \texttt{transformed}\_d_i^{0.98}]
  \end{equation}
Cross fold validation with 10 folds is used in MLR model 2. From the model summary in Table-\ref{table:MLR 2}, it is observed that the 10-fold cross validation score is 81.15\% and the adjusted R2 score is 83.11\%, demonstrating little overfitting. Table-\ref{table:MLR 2} also represents an ANOVA table where the transformed $d_i^{0.98}$ contributes 47.39\% to the model and $d_i^{0.99}$ contributes 36.01\% to the model. None of the p-values are greater than 0.05. From coefficients, notice that none of the VIF values are greater than 10, meaning there is no multicollinearity. Figure-\ref{fig:Original vs Predicted Observation in MLR model 2} shows the relationship between the original transformed $d_i^{0.80}$ and the values predicted by the MLR model. There is a correlation of 0.91 between original and predicted value. This model is well fitted but there is an outlier observable. The variance not explained by the model may be due to this outlier. One can remove that outlier but doing so makes the model less viable. 

\subsection{Residual Analysis for MLR model 2}

\begin{figure}[t]
  \centering
  \subfloat[Residual plot for MLR model 2]{\includegraphics[width=0.45\textwidth]{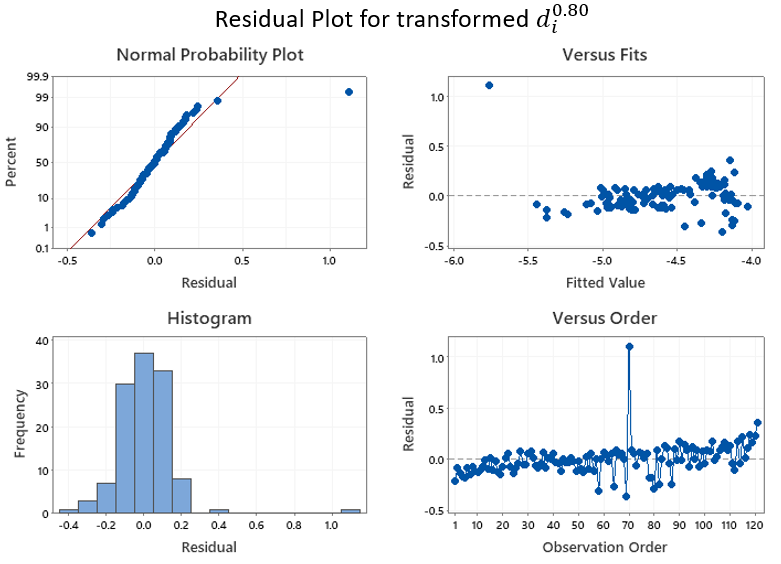}}
  \hfill
  \subfloat[Residuals vs $d_i^{0.99}$ plot for MLR model 2]{\includegraphics[width=0.45\textwidth]{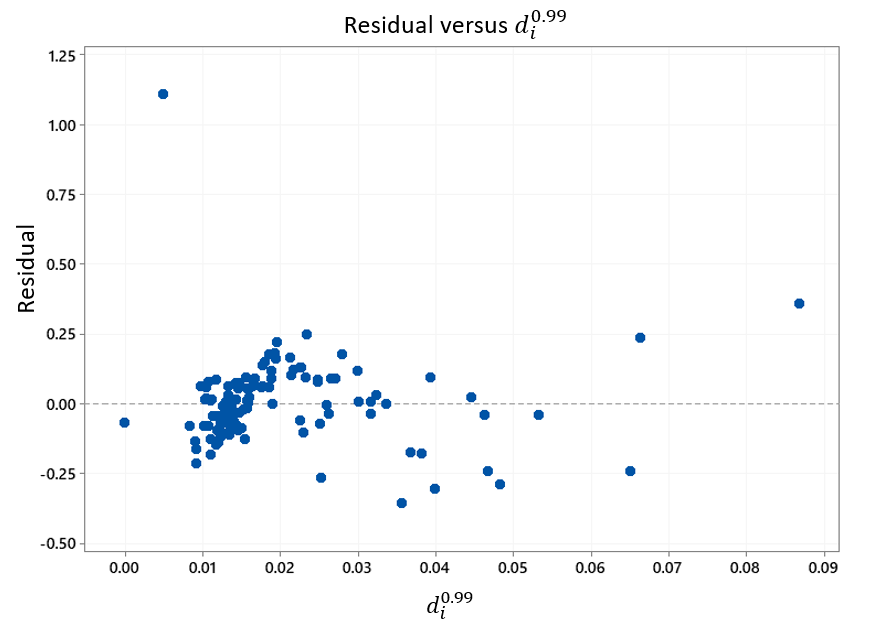}\label{fig:res2}} \\
  \subfloat[Residuals vs transformed $d_i^{0.98}$ plot for MLR model 2]{\includegraphics[width=0.45\textwidth]{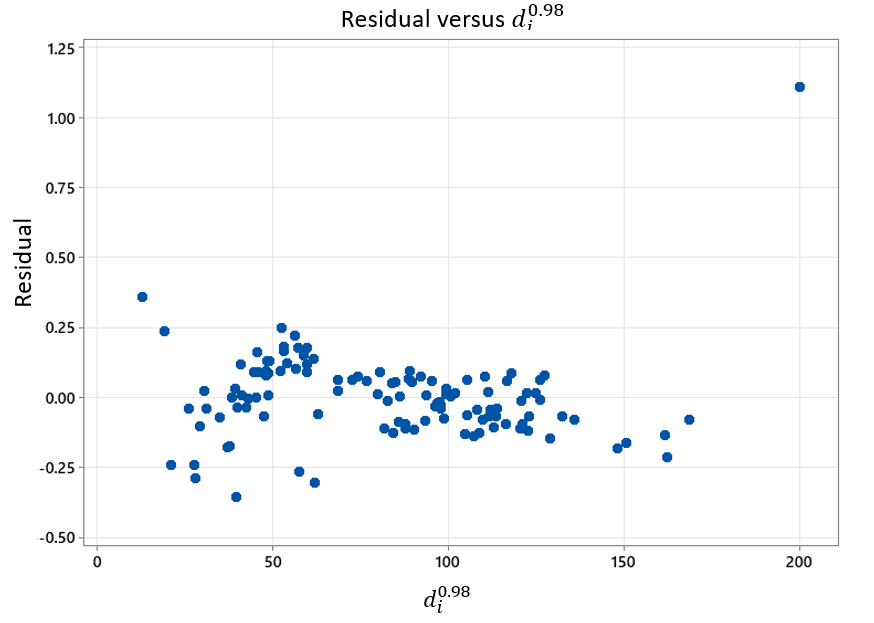}\label{fig:res3}}
  \caption{Residual Analysis for MLR 2}
  \label{fig:residual analysis}
\end{figure}

For the MLR model, some of the assumptions are that residuals should be normally distributed and there should be homogeneity of variance. The residuals histogram and normal probability plot for model 2 can be found from Figure-\ref{fig:residual analysis} which shows that the residuals are almost normal with exception of one outlier. Residual relation with fitted value and observation order also can be obtained from the same figure indicating that the residual has equal variance except one outlier. Figure-\ref{fig:res2} and Figure-\ref{fig:res3} represents the predictor vs residual plot for two selected predictors. There is also one unusual observation seen in that plot. The Durbin-Watson statistics value is 1.86; this value is close to 2 which demonstrates that there is no residual correlation. Further investigation should be done for finding the unusual observation and find out if there is any potential option to remove it.

\subsection{Result of Decision Tree Model}

The decision tree model obtained training and testing accuracies of 99.73\% and 97.54\% accordingly. Correlations between original and predicted values are 0.99. This model is overfit to training data.

\subsection{Result of RFR Model}

\begin{figure}[!tb]
  \centering
  \subfloat[For training samples]{\includegraphics[width=0.45\textwidth]{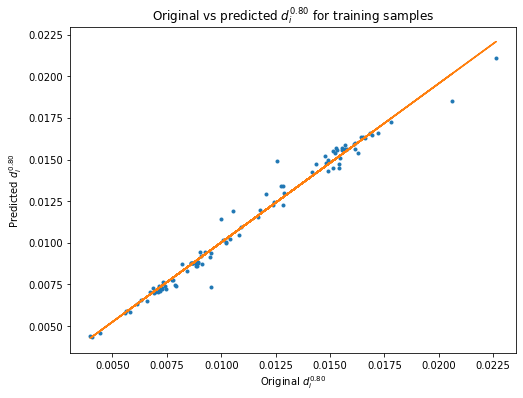}\label{fig:rfr1}}
  \hfill
  \subfloat[For testing samples]{\includegraphics[width=0.45\textwidth]{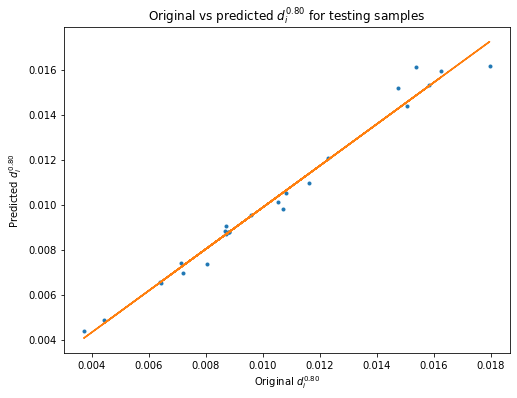}\label{fig:rfr2}}
  \caption{Original vs Predicted $d_i^{0.8}$ for training and testing data in RFR model}
  \label{fig:original vs predict for rfr}
\end{figure}

\begin{table}[!tbp]
 \caption{PERFORMANCE OF RFR MODEL}
  \vspace{2pt}
    \centering
    \begin{tabular}{ccccc}
    \hline
        RFR & Correlation & $R^2$ value & MAPE & RMSE \\ \hline
        Traning Sample & 0.99 & 0.98 & 3.21 & 0.0005 \\
        Testing Sample & 0.99 & 0.97 & 4.22 & 0.0005 \\ \hline
    \end{tabular}
    \label{table:PERFORMANCE OF RFR MODEL}
\end{table}

\begin{figure}[!tbp]
    \centering
    \includegraphics[width= 10cm]{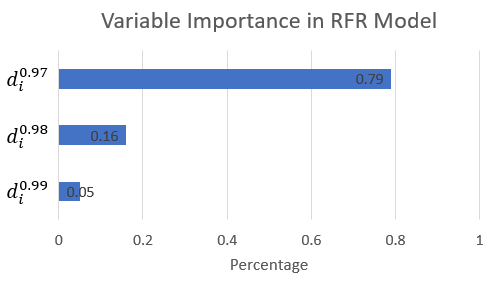}
    \caption{Variable importance in RFR model}
    \label{fig:variable importance}
\end{figure}

Table-\ref{table:PERFORMANCE OF RFR MODEL} represents the results of the RFR model. This model is giving 98.02\% training accuracy and 97.73\% testing accuracy which validates the model with no overfitting. Error is calculated through mean average percentage error (MAPE) and root mean square error (RMSE). For this analysis, MAPE will be effective as our feature values are so small. In this method, the MAPE value for training samples is 3.21\% and 4.22\% for testing samples. Original vs predicted $d_i^{0.80}$ plots for training and testing samples are shown in Figure-\ref{fig:original vs predict for rfr}. The RFR model fits both training and testing samples well with 0.99 correlation value. One can find the variable importance from this model which is shown in Figure-\ref{fig:variable importance}. For this model, $d_i^{0.97}$ has the highest importance with 79\%, followed by $d_i^{0.98}$ with 16\% importance, and lastly $d_i^{0.99}$ with 5\% importance. This RFR model utilized all three of the variables without any transformation.

\section{Comparison of results and discussion}
\begin{table}[t]
 \caption{PERFORMANCE COMPARISON OF FOUR MODEL}
  \vspace{2pt}
    \centering
   \begin{tabular}{m{0.1\linewidth} m{0.1\linewidth} m{0.1\linewidth} m{0.1\linewidth} m{0.1\linewidth} m{0.1\linewidth} m{0.1\linewidth}}
    \hline
        Model & Training Accuracy & K-fold/ Testing Accuracy & Correlation Coefficient & Model Overfit & Multi-collinearity problem &  Outlier Problem \\ \hline \vspace{1pt}
        MLR $1$ & $77.30\%$ & $70.52\%$ & $0.88$ & Yes & Yes & No \\
        MLR 2 & 83.11\% & 81.15\% & 0.91 & Yes & No & Yes \\
        Decision Tree & 99.73\% & 97.54\% & 0.99 & Yes & No & No \\
        RFR & 98.02\% & 97.73\% & 0.99 & No & No & No \\ \hline
    \end{tabular}
    \label{table:comparison}
\end{table}

\begin{figure}[!tbp]
    \centering
    \includegraphics[width= 10cm]{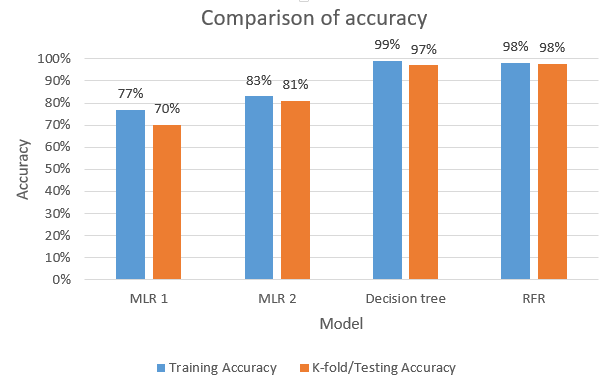}
    \caption{Accuracy comparison among four models}
    \label{fig:Accuracy comparison among four models}
\end{figure}

An overall comparison of multiple linear regression models versus tree-based regression models is presented in Table-\ref{table:comparison} and Figure-\ref{fig:Accuracy comparison among four models}. MLR model 1 is built without data transformation, suffers from overfitting and multicollinearity and achieves a comparatively poor accuracy. On the other hand, MLR model 2 utilizes data transformation and stepwise regression, which leads to a model with reduced overfitting and no multicollinearity. While MLR 2 has improved accuracy compared to MLR 1, it is still significantly worse than tree-based models. Both MLR models contain predictive outliers. The decision tree model obtains very good accuracy (97.5\%) but is overfit. The RFR model is an ensemble method and therefore retains the accuracy of the decision tree model (and slightly improves upon it) while significantly reducing overfitting. Furthermore, there are no outliers in either tree-based model.\\ 

The tree-based models are not negatively impacted by the multicollinearity and non-normality inherent in this dataset, making them an ideal choice for datasets with skewed distributions and multicollinear features. Tree-based models are non-parametric and able to cope with the highly non-linear effect of the features. Residual analysis is done for the MLR model for justifying the residual assumption of MLR which is not needed for the RFR model. The correlation coefficient between original and predicted value is higher in tree-based models. For the best tree-based model (RFR), $d_i^{0.97}$ turns out to be the most important predictor where the best MLR model (model 2) finds transformed $d_i^{0.98}$ as the most valuable predictor.\\ 

There is a very good linear relationship between $d_i^{0.80}$ and $n_i^{0.8}$(cycle number at EOL). As it is now possible to predict $d_i^{0.80}$ from the early cycle b values with the help of tree-based regression very precisely, then it will be feasible to get the battery cycle life from that predicted $d_i^{0.80}$. In this study, all the early cycle $d$ values is utilized to predict $d_i^{0.80}$. In future study, combination of independent features (early cycle $d$ values) will be used to predict battery cycle life.\\

However, this dataset is limited to only 123 battery cell and low in high cycle life battery cell. If it is possible to accumulate more battery cell in the dataset and reduce the skewness of the features, the predictive model will be more robust.

\section{Conclusion}

From this study, one can conclude that when a dataset exhibits multicollinearity and is highly skewed or non-normal, it is beneficial to use a model, such as tree-based regression, which is non-parametric and non-linear. The RFR model is fairly robust and outperforms all other models in terms of accuracy (achieving 97.7\% testing accuracy). The RFR model uses all the features and offers insight into the relative importance of them. If linear regression is to be used, this study demonstrates the benefit of transforming the data and utilizing stepwise regression to address multicollinearity and lack of normality. Interestingly, in our study the tree-based models also exhibited no predictive outliers, whereas the linear regression models did.

\section*{Acknowledgement}
We would like to thank K. A. Severson, P. M. Attia, William Chueh, and their co-authors for their generously providing the data used in their study for the use in this work. This work was supported under Idaho National Laboratory’s Laboratory Directed Research and Development program (LDRD No. 19P45-013FP). The INL is operated by Battelle Energy Alliance under Contract No. DE-AC07-05ID14517 for the U.S. Department of Energy. The U.S. Government retains and the publisher, by accepting the article for publication, acknowledges that the United States Government retains a nonexclusive, paid-up, irrevocable, world-wide license to publish or reproduce the published form of this manuscript, or allow others to do so, for U.S. Government purposes.

\bibliography{references}

\end{document}